# A Case Study on the Independence of Speech Emotion Recognition in Bangla and English Languages using Language-Independent Prosodic Features

Fardin Saad, Hasan Mahmud, Mohammad Ridwan Kabir, Md. Alamin Shaheen, Paresha Farastu, Md. Kamrul Hasan

**Abstract:** A language agnostic approach to recognizing emotions from speech remains an incomplete and challenging task. In this paper, we performed a step-by-step comparative analysis of Speech Emotion Recognition (SER) using Bangla and English languages to assess whether distinguishing emotions from speech is independent of language. Six emotions were categorized for this study, such as - *happy*, *angry*, *neutral*, *sad*, *disgust*, and *fear*. We employed three Emotional Speech Sets (ESS), of which the first two were developed by native Bengali speakers in Bangla and English languages separately. The third was a subset of the Toronto Emotional Speech Set (TESS), which was developed by native English speakers from Canada. We carefully selected language-independent prosodic features, adopted a Support Vector Machine (SVM) model, and conducted three experiments to carry out our proposition. In the first experiment, we measured the performance of the three speech sets individually, followed by the second experiment, where different ESS pairs were integrated to analyze the impact on SER. Finally, we measured the recognition rate by training and testing the model with different speech sets in the third experiment. Although this study reveals that SER in Bangla and English languages is mostly language-independent, some disparities were observed while recognizing emotional states like *disgust* and *fear* in these two languages. Moreover, our investigations revealed that non-native speakers convey emotions through speech, much like expressing themselves in their native tongue.

## 1 Introduction

Humans have been intuitively using speech as the most natural and preferred means of communication. However, speech signals are mostly non-stationary processes, containing multiple components that vary in time and frequency. Since these signals occur naturally, they are erratic in nature [1]. Furthermore, these signals carry a lot of information and, at the same time, convey an individual's emotional status. An emotional speech expresses the patterns of rhythm and intonation, often referred to as prosody in a language [2], [3]. The prosodic cues in speech signals are also known as para-linguistic features since they are linked with speech segments properties such as syllables, words, and sentences [3]. Zeng et al. [4] demonstrated prosodic features to contain the most critical and exclusive emotional information. Consequently, numerous research in Speech Emotion Recognition (SER) have used prosodic features to identify emotional states [3].

SER directs the identification of emotional states of a person from their speech [5]. It facilitates the measurement of acoustic cues from speech as a standard for emotion recognition [6], which led the path for researching the most comprehensive features that contribute greatly to this cause. In line with this, Pell et al. [7] emphasized the significance of vocal features for emotion recognition from speech. Several studies were conducted to investigate the salient vocal features. Amongst them, prosodic features were found to give better if not the same emotion recognition accuracy than human judges [3], [8], [9]. Apart from this, SER systems are useful in several applications such as e-learning, where the emotional attributes of the pupils can be identified to regulate the teaching style of the tutors [10], [11]. Rapid commercialization of speech emotion recognition can be seen in employee mood identification [12], interactive games [13], call centers [14] to decipher customer queries and complaints, psychiatric aids, among others.

Feature extraction and selection from speech impart great importance in successfully identifying emotional states. However, selecting a large number of features breeds various complexities, which eventually results in classification error [3], [15]. Furthermore, languages can differ in many ways with regards to their grammatical and morphological properties [15], making SER ambiguous. The variations in dialects of a language can occur due to the usage of words, accents, or how people arrange their speech. These variations can be credited to certain social factors, culture, or geographical distance [16]. Moreover, factors such as – gender, age, background, vocal features, etc., of a speaker highly influence the expression of different emotions [2], [17], [18]. Due to these multifarious factors, emotion recognition from speech remains an arduous task [19]. Hence, there may be particularly pivotal factors that influence speech emotion recognition [20]. So, it becomes essential for us to work with vocal features which are independent of the nuances of language.

From previous research, we found that prosodic features [3], [4], [21] such as fundamental frequency or pitch-related features containing pitch mean, pitch median, pitch standard deviation, and energy-related features like intensity carry a lot of emotional information [22], [23]. In contrast, the spectral features such as MFCC related features depend on phonemes, and thus the style of an utterance [23]. Therefore, it can be inferred that spectral features are language-dependent [23] features, whereas prosodic features are language-independent [21], [22].

Schull et al. [24] enforced wrapper-based search along with the dynamic base contour to develop a feature set from intensity, pitch, formants, and MFCCs. The best emotion recognition rate has been accomplished using MFCC features with SVM. Rajoo et al. [20] used MFCCs, formants, energy, and fundamental frequency or pitch as acoustics cues for speech emotion recognition. Borchert et al. [3], [25] used prosodic vocal features and quality features such as formants, spectral energy distribution, jitter, and shimmer, to classify emotions. For deducing the salient set of vocal features, Kostoulas et al. [26] and Anagnostopoulos et al. [27] employed a subset of correlated features. Despite improving the emotion classification rate, these features are not language-independent since most used spectral features on a single emotional corpus.

Thus, to construct a Speech Emotion Recognition system, it is vital to select prosodic features such as fundamental frequency or pitch and energy-related features and select a classifier independent of these features. Noroozi et al. [15] presented a system for achieving a set of language-independent features. They observed that pitch and energy-related features were language-independent when they adopted a Support Vector Machine (SVM) classifier. While exploring language-independent features via feature selection, Shaukat et al. [22] discerned that fundamental frequency or pitch, formants, intensity or energy, harmonicity, loudness, duration, etc., were language-independent features. Subsequently, they achieved a higher performance rate when using SVM for their classification method. Lieke et al. [28] used 3 prosodic cues onto speech by Spanish learners of Dutch to judge whether it improves native Dutch speakers' perception of accentedness and comprehensibility. Luengo et al. [8] illustrated that six prosodic features incorporated with an SVM classifier could achieve an emotion recognition rate almost equal to GMM models with 512 spectral MFCC features emphasizing the significance of prosodic features in distinguishing emotions. Rao et al. [29] extracted and employed local and global prosodic features with SVM at different speech segments to recognize emotions from the Telegu emotional speech corpus. Bhatti et al. [21] used 17 prosodic features, which included pitch mean, pitch median, pitch standard deviation, etc., as language-independent features for identifying emotional states from speech.

In this study, we aim to juxtapose and compare Bangla and English languages to measure the language independency of SER using language-independent prosodic features and native/non-native speakers. We used 6 emotional states such as - *happy, angry, neutral, sad, disgust,* and *fear* [1], [2], [20], [23], in English and Bangla languages across three Emotional Speech Sets (ESS), such as – 1) the English TESS (E-TESS), a subset of the Toronto ESS (TESS) [30], for the native English language, 2) the Bangla ESS (BESS) for the native Bangla language, developed by native Bangla speakers in accordance with the development process of TESS [31], and 3) similar to BESS, the English ESS (EESS) for the non-native English language, developed by the same native Bangla speakers, or in other words non-native English speakers. We constructed a feature set through careful selection of language-independent features such as pitch mean, pitch median, pitch standard deviation and intensity [1], [15], [21]–[23] and analyzed its influence on SER. These 4 features are the most commonly used and salient prosodic features in almost all the SER systems that used prosodic features with SVM classifier [1], [3], [8], [9], [15], [21]–[23]. Furthermore, we avoided choosing too many features to preclude misclassification, correlated features, and overlap. Since our objective is not to increase the performance of our SER system instead to analyze the nuances in Bangla and English languages, we adopted SVM [32] for classifying the emotions mentioned earlier. However, we also employed traditional classifiers for SER, such as – Hidden Markov Model (HMM), Gaussian Mixture Model (GMM), Artificial Neural Networks (ANN), and k-Nearest Neighbor (kNN) [3] only to compare the emotion recognition rate of these classifiers with SVM.

In Section 2, we thoroughly discussed about our proposed approach, followed by feature extraction and selection, in Sections 3. In the subsequent sections, we have elaborated on our classification approach (Section 4), investigated the language independence of Speech Emotion Recognition for Bangla and English languages (Section 5), and discussed its scope and future possibilities (Section 6).

## 2 Emotional Speech Set Acquisition and Development

Due to the lack of any significant emotional speech corpora for Bangla language, we have developed a Bangla ESS (BESS), in accordance with the Toronto ESS (TESS) [30], as shown in Fig. 1, with assistance from 11 native Bangla speakers (64%

Male, 36% Female) who are undergraduate students of the affiliated institution. Similarly, the same native speakers, or in other words, the same non-native English speakers, participated in the development of the English ESS (EESS). The 11 native Bangla speakers were carefully selected for this study based on their fluency and proficiency in both Bangla and English languages.

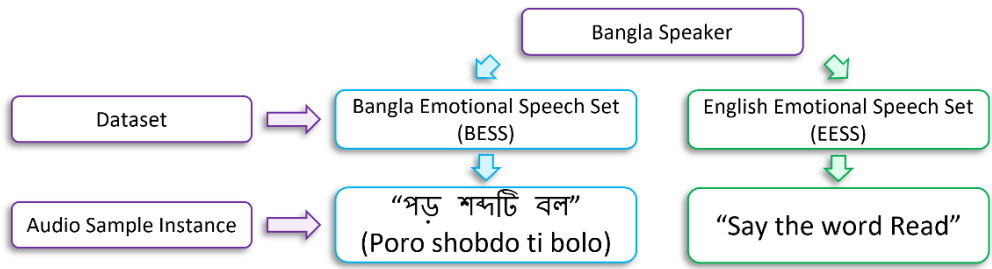

Fig. 1. Construction of Bangla Emotional Speech Set (BESS) and English Emotional Speech Set (EESS) datasets by native speakers fluent and proficient in both Bangla and English.

We identified 6 of the most common emotional states used in SER, such as – *happy*, *angry*, *neutral*, *sad*, *disgust*, and *fear* for our speech corpora [2], [15], [20], [23], [31]. The 11 speakers were asked to elicit emotional speech, simulating these emotions. Approximately 30-40 hours were spent for recording each emotional state, among which 5-6 hours were allotted for rehearsal and calibration of the speaker's portrayal of each emotion. At least one male and one female speaker were involved in recording the audio samples of each emotion.

Each audio sample, containing the phrase, "Say the word…", followed by a monosyllabic noun [30], [31], was recorded within 1-2 seconds. For instance, it can be observed from Fig. 1 that a sentence, used in the dataset was, "Say the word **Read**." The monosyllabic nouns were selected based on the existing speech intelligibility such as the Northwestern University Auditory Test-Number 6 (NU-6) [33]. While developing the Bangla dataset, the same sentence, "Say the word Read" in English, was translated to Bangla, which read as, "Poro shobdo ti bolo", as shown in Fig. 1. It can be discerned from Fig. 2 that the verbatim translation of Bangla words such as – "Poro", "shobdo ti", and "bolo" to its corresponding English words are "Read", "the word", and "Say", respectively. For comprehensive understanding, Bangla words used in the Bangla speech sample were mapped to their corresponding English words, and in this manner, BESS and EESS were developed.

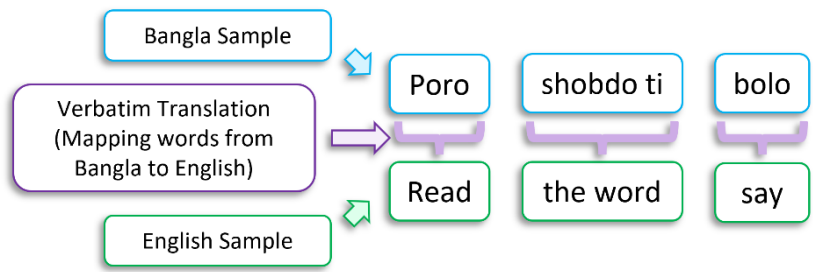

Fig. 2. Verbatim translation of Bangla words to the corresponding English words.

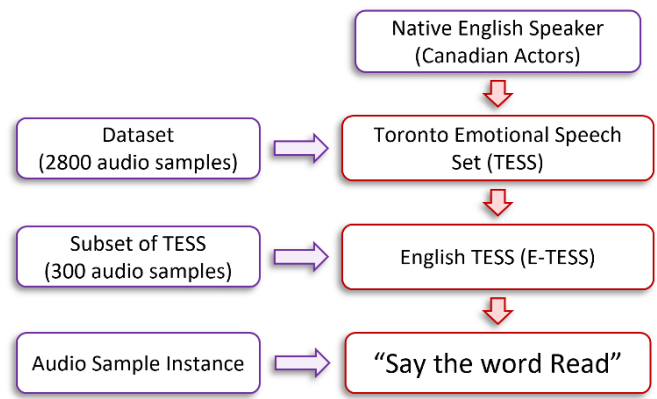

Fig. 3. Formation of the English TESS (E-TESS) dataset as a subset of the existing Toronto Emotional Speech Set (TESS).

Although the Auditory Test-Number 6 (NU-6) housed 200 words [31], we carefully selected 50 words whose Bangla translation would adhere to the lexical and semantic properties of speech intelligibility. Therefore, for the 2 datasets (BESS and EESS) and for the 6 emotions (*happy*, *angry*, *neutral*, *sad*, *disgust*, and *fear*) with 50 audio speech samples each, we have a total of 2 datasets × 6 emotions × 50 speech samples = 600 audio samples, 300 samples per dataset (BESS and EESS).

On the other hand, TESS was developed by 2 Canadian actors, of whom one was younger (26 years of age), and the other was older (64 years of age), containing a total of 2800 audio samples (200 NU-6 words x 7 emotions x 2 actors) [31]. However, as established earlier, to ensure preservation of the speech intelligibility for Bangla and English languages, audio samples containing the same 50 words as BESS or EESS, were retrieved from TESS for the 6 emotions to create the dataset, English TESS (E-TESS) as a subset of TESS, containing 300 samples. The workflow of generating E-TESS is depicted in Fig. 3. In this manner, 3 datasets were assembled for our experiments, which will be addressed as Bangla Emotional Speech Set (BESS), English Emotional Speech Set (EESS), and English TESS (E-TESS), as shown in Fig. 1 and Fig. 3.

## 3 Feature Extraction and Selection

Features dictate the performance of a Speech Emotion Recognition system, and identifying emotional states require various features [20]. For the expression of emotions, speech features contribute in particular ways [34]. However, the most salient features can improve the performance of a model exponentially. Misclassification and overlaps of emotional states by an SER system are mostly due to inadequate and ineffective choices of features [35]. In our work, the vocal features for both BESS and EESS were extracted using signal processing techniques, implemented in the open-source software package, "*Praat*", developed by the University of Amsterdam [36]. This process is briefly elaborated on in the following section. For this study, we used language-independent prosodic features, such as – *pitch median, pitch mean, pitch standard deviation*, and *intensity*.

- Pitch or fundamental frequency [37], [38] is frequently deployed in Speech Emotion Recognition Systems [3], [15], [20], [39], [40]. It represents the vibration frequency of the vocal cords during sound production.
- Pitch-related features such as pitch mean and pitch median are widely used in SER systems [1], [21], [23] and are regarded as language-independent features [15], [21]–[23].
- Pitch standard deviation is another prosodic feature which carries vital emotional information [21], [23] and is used as one of the acoustic feature [21], [41].
- Energy or intensity refers to the loudness of sound. This feature is also used quite frequently in SER systems [16], [42]. It is deemed to be a language-independent prosodic feature [15], [22].

## 3.1 Speech Normalization and Feature Analysis using Praat Software

The audio samples were scrupulously recorded with a high-quality microphone at a sampling rate of 44 kHz, or in other words 44100 samples per second per channel [36] were recorded for maximum audio quality. A quiet room was selected for the recording sessions to reduce background noise. No filters were applied on the audio recordings to avoid distortion and to ensure preservation of the original intensity of the speech signals.

The sample audio recordings of BESS and EESS for the phrases "Poro shobdo ti bolo" and "Say the word Read", eliciting the emotional state, *happy*, are illustrated in Fig. 4a and Fig.4b, respectively. As demonstrated in Fig. 2, these phrases are translations of each other.

In each of Fig. 4a and Fig. 4b, the audio signals are divided into 2 parts, such as – 1) the *time-domain* audio signal, depicted in the upper portion of each figure, with the x-axis representing *time* and the y-axis representing the *amplitude* of the signal, and 2) a *frequency* vs *time* plot of the corresponding audio signal, depicted in the lower portion of each figure [36]. If observed carefully, the *amplitude* of the Bangla audio signal in Fig. 4a, varies compared to that of the English audio signal in Fig. 4b. The peaks of amplitude for the Bangla audio sample can be segregated into individual words such as "Poro", "shobdo", "ti", and "bolo". Similarly, the English audio sample in Fig. 4b, can be separated into individual words such as "Say", "the", "word", and "Read". Upon further inspection and keeping in mind that both audio samples were recorded for the *happy* emotion, it can be discerned that there are intermittent signal gaps for the Bangla audio sample (Fig. 4a). However, such manifestation cannot be inferred for the English audio sample (Fig. 4b).

Essentially, in the English language, the sentence pattern is set as – the *subject*, then the *verb*, and finally the *object* [43] whereas, in Bangla, it follows the pattern – the *subject*, then the *object*, and finally the *verb* [44]. Furthermore, from the verbatim translation of Bangla to English words in Fig. 2, it is evident that both languages adhere to a different sentence pattern. The variation in the speech signals for English and Bangla languages is mainly due to these nuances. However, as mentioned earlier, since we are only using prosodic features, which are language-independent [21], [22], the style of utterance does not matter, and these nuances do not hamper the classification of emotions from varying speech signals.

Considering the two prosodic features, *pitch standard deviation* and *intensity* for the emotional states, *happy* and *disgust*, 2 feature distribution plots, combining these features for all the 3 datasets (BESS, EESS, and E-TESS), are depicted in Fig. 5a and Fig. 5b, respectively. In the feature distribution plot for the *happy* emotion (Fig. 5a), the features for the 3 datasets appear to be clustered together. This observation further clarifies that the prosodic features are independent of the nuances of language, and therefore, it may be stated, "*for Bangla and English languages, the emotional state 'happy' is language and speaker (native/non-native) independent*". On the other hand, from the feature distribution plot of the "*disgust*" emotion (Fig. 5b), the two prosodic features (*pitch standard deviation* and *intensity*) for the E-TESS dataset do not overlap with those of the BESS and the EESS datasets whereas, they appear to be overlapping for latter two datasets. Therefore, from this illustration, it may be surmised, "*for Bangla and English languages, the emotional state 'disgust' is language and speaker (native/non-native) dependent*". The validity of these two statements will be further investigated in Section 5.

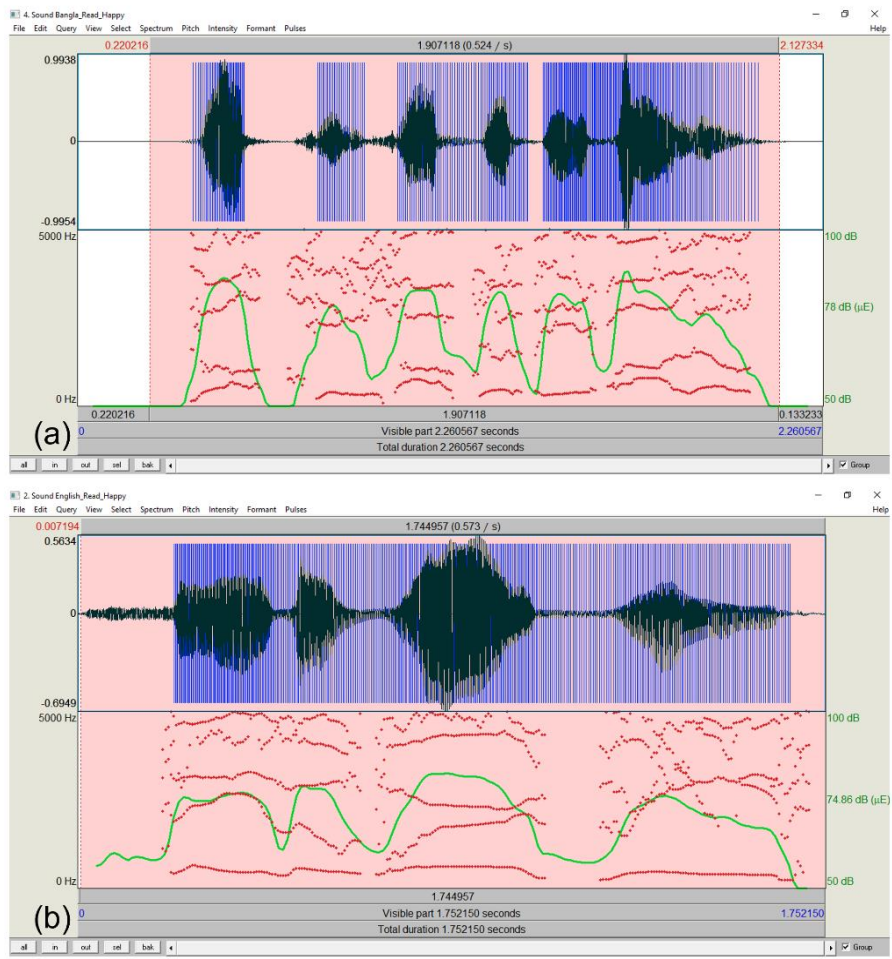

Fig. 4. Audio samples for "*happy*" Emotion for – (a) the phrase "Poro shobdo ti bolo" of the Bangla Emotional Speech Set (BESS) (b) the phrase "Say the word Read" of the English Emotional Speech Set (EESS).

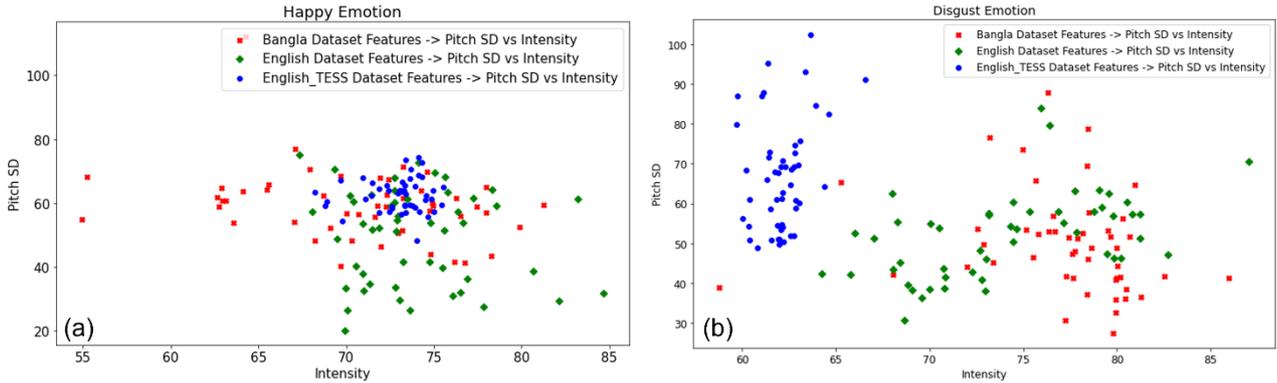

Fig. 5. Feature (Pitch Standard deviation vs Intensity) distribution plot of the datasets BESS, EESS, and E-TESS for – (a) the *happy* emotion (b) the *disgust* emotion.

## 4 Classifier

Support Vector Machine (SVM) [32] is one of the conventional classifiers deployed in SER Systems [3], [6], [18], [34], [39], [45]–[48]. The models that deploy SVM classifier, designate new training examples to any one category, making this binary linear classifier, a non-probabilistic one, in the process. Using a phenomenon known as kernel trick, SVMs can effectively execute non-linear classifications alongside linear ones. As mentioned earlier, the language-independent vocal features [21] that are selected for SER, have a better recognition rate when SVM is used [15], [22]. Conventionally known as a separating hyperplane, a Support Vector Machine may be denoted as a discriminative classifier.

Nevertheless, we used various traditional SER classifiers, such as – HMM, GMM, ANN, and kNN [3] for evaluating their performance on the 3 datasets (BESS, EESS, E-TESS). As depicted in Table 1, the average emotion classification rate using an SVM classifier supersedes the other traditional classifiers (HMM, GMM, kNN, and ANN) of SER in all the datasets. Therefore, in this study, we adopted a generic SVM kernel analogous to the classifiers deployed in SER to generate a non-linear hyperplane for recognizing the 6 emotional states (*happy*, *angry*, *neutral*, *sad*, *disgust*, and *fear*). Additionally, we avoided using any deep learning-based classifiers due to their overfitting complexities and preliminary processing of the ESSs.

Table 1: Recognition Rate of different SER classifiers for Bangla (BESS), English (EESS) and English TESS (E-TESS) datasets.

| Emotional Speech Set (ESS) | Emotion Recognition Rate (%) | | | | |
|---|---|---|---|---|---|
| | HMM | GMM | kNN | ANN | SVM |
| Bangla Emotional Speech Set (BESS) | 45 | 46.7 | 81.3 | 68.3 | 88.3 |
| English Emotional Speech Set (EESS) | 38.3 | 42 | 75 | 71.7 | 85 |
| English TESS (E-TESS) | 58.3 | 56.7 | 93.3 | 86.6 | 93.3 |

## 5 Experiments and Result Analysis

As mentioned earlier, each of the 3 datasets, namely, BESS, EESS, and E-TESS had a total of 300 audio speech samples, i.e., 50 audio samples for each of the 6 emotions. Considering these datasets, 3 types of experiments were conducted for investigating the language independent nature of SER, and eventually unearthing other disparities corresponding to emotions and native/non-native speakers, such as – 1) on individual speech sets, 2) on integrated speech sets, and 3) by training the model using one speech set and testing it with a different speech set. We elaborate on these 3 experiments in the following sub-sections.

### 5.1 Experiment 1: Individual Speech Set

In this experiment, the SVM classifier was trained and tested on each of the 3 datasets individually, with an 80-20, train to test split of the 300 audio samples (240 training and 60 testing samples). The overall emotion recognition rate across the datasets BESS, EESS, and E-TESS were recorded to be 88.3%, 85%, and 93.3%, respectively. The average performance of the 6 emotional *states happy, angry, neutral, sad, disgust*, and *fear* were 83.3%, 90%, 96.7%, 93.3%. 86.7%, and 83.3%, respectively. From this result, it can be inferred that SER works well only when one language is involved. However, for EESS, the model has a lower recognition rate compared to E-TESS. The recognition rates of the SVM model for individual ESS and different emotions are summarized in Table 2. It is evident from the literature [20] that when using a second language, speakers are prone to feel less strongly due to fewer recollections and deep-rooted memories. Therefore, a probable reason for a lower recognition rate of the model for EESS may be attributed to the fact that this dataset was developed by the 11 Bangle native speakers, who despite being fluent in English, could not express their emotions in English as effectively as they could have in Bangla.

Table 2: Experiment 1 – Recognition rate of SVM for the individual Emotional Speech Set (ESS).

| Emotional Speech Set (ESS) | Overall Recognition Rate (%) | Emotion Recognition Rate (%) | | | | | |
|---|---|---|---|---|---|---|---|
| | | *happy* | *angry* | *neutral* | *sad* | *disgust* | *fear* |
| Bangla Emotional Speech Set (BESS) | **88.3** | 90 | 100 | 90 | 90 | 90 | 70 |
| English Emotional Speech Set (EESS) | **85** | 70 | 80 | 100 | 100 | 80 | 80 |
| English TESS (E-TESS) | **93.3** | 90 | 90 | 100 | 90 | 90 | 100 |
| Overall Performance | **88.87** | 83.3 | 90 | 96.7 | 93.3 | 86.7 | 83.3 |

## 5.2 Experiment 2: Integrated Speech Set

In this experiment, 2 different Integrated ESS (IESS) were formed by concatenating the audio samples of one dataset with another, such as – 1) IESS-1, formed by integrating BESS and EESS, and 2) IESS-2, formed by integrating BESS and E-TESS. Each IESS had a total of 600 audio samples, which at 80-20, train to test split ratio (480 training and 120 testing samples) were used for training and testing the SVM classifier.

For IESS-1, with BESS and EESS, having individual recognition rate of 88.3% and 81.7%, respectively, an overall accuracy of 85% was observed. On the other hand, IESS-2 had an overall classification rate of 83.3% while each of the datasets, BESS and E-TESS, had a performance rate of 75% and 91.7%, respectively. However, the classification rate of this experiment decreased with respect to the first one. The recognition rates of the SVM classifier for the IESS-1 and IESS-2, are summarized in Table 3 and Table 4, respectively.

Since both datasets, BESS and EESS, were developed by native Bangla speakers, the higher overall emotion recognition rate for IESS-1 compared to that for IESS-2, perhaps suggests that non-native speakers tend to express their emotions in English, likewise their native tongue. Furthermore, the higher recognition rate for BESS compared to that for EESS may be attributed to the native speaker's inability to naturally express their emotions in English [20]. Again, since the datasets, BESS and E-TESS, were developed by Bangla and English native speakers, respectively, higher recognition rate for E-TESS compared to that for BESS leads us to believe that there may be certain differences in language, governing how different emotions are expressed.

Table 3: Recognition Rate of SVM for the Integrated Emotional Speech Set – 1 (IESS-1), consisting of BESS and EESS.

| Emotional Speech Set (ESS) | Recognition Rate (%) | Emotion Performance (%) | | | | | |
|---|---|---|---|---|---|---|---|
| | | *happy* | *angry* | *neutral* | *sad* | *disgust* | *fear* |
| Only Bangla (BESS) | **88.3** | 90 | 100 | 80 | 100 | 90 | 70 |
| Only English (EESS) | **81.7** | 90 | 80 | 90 | 90 | 70 | 70 |
| Bangla & English (IESS-1) | **85** | 90 | 90 | 85 | 95 | 80 | 70 |

Table 4: Recognition rate of SVM for the Integrated Emotional Speech Set – 2 (IESS-2), consisting of BESS and E-TESS.

| Emotional Speech Set (ESS) | Recognition Rate (%) | Emotion Performance (%) | | | | | |
|---|---|---|---|---|---|---|---|
| | | *happy* | *angry* | *neutral* | *sad* | *disgust* | *fear* |
| Only Bangla (BESS) | **75** | 50 | 100 | 80 | 90 | 70 | 60 |
| Only English TESS (E-TESS) | **91.7** | 100 | 80 | 100 | 70 | 100 | 100 |
| Bangla & English TESS (IESS-2) | **83.3** | 75 | 90 | 90 | 80 | 85 | 80 |

## 5.3 Experiment 3: Distinct Speech Set for Training and Testing

Considering the 3 datasets (BESS, EESS, E-TESS), the final experiment involved training the SVM classifier by one speech set and separately testing it with the remaining two. For instance, as illustrated in Table 5, the model was firstly trained using BESS, followed by separately testing it with the remaining datasets (EESS and E-TESS). Similarly, other combinations of the datasets were considered, and emotion recognition rates were recorded.

With BESS as the training dataset, the model achieved recognition rates of 76.7% and 55%, for EESS and E-TESS datasets, respectively. However, with E-TESS as the training dataset, the recognition rate of the model for both BESS and EESS was 45%.

As observed from Table 5, the classification rate of this experiment decreased even further compared to the previous two, which may be credited to the failure of recognizing the emotional states, "*disgust*" and "*fear*" at all, for the following combinations – 1) BESS as training, E-TESS as testing, 2) E-TESS as training, EESS as testing, and 3) E-TESS as training, BESS as testing. These findings lead us to believe that the prosodic cues for the emotions, "*disgust*" and "*fear*", may be different in Bangla and English when expressed by their respective native speakers. Furthermore, there might be a possibility that these emotions are influenced by the culture or background of the speaker [2], [15]. However, the remaining emotional states were moderately recognized for this pair of training and testing datasets.

Table 5: Experiment 3 – Recognition rate of SVM when using distinct Emotional Speech Set for training and testing.

| Training Dataset | Testing Dataset | Recognition Rate (%) | Emotion Performance (%) | | | | | |
|---|---|---|---|---|---|---|---|---|
| | | | *happy* | *angry* | *neutral* | *sad* | *disgust* | *fear* |
| Bangla (BESS) | English (EESS) | **76.7** | 60 | 90 | 80 | 90 | 60 | 80 |
| | English TESS (E-TESS) | **55** | 100 | 80 | 80 | 70 | **0** | **0** |
| English TESS (E-TESS) | English (EESS) | **45** | 50 | 100 | 50 | 70 | **0** | **0** |
| | Bangla (BESS) | **45** | 50 | 100 | 40 | 80 | **0** | **0** |

On the other hand, the recognition rate of the model for EESS was recorded to be comparatively higher with BESS as the training set compared to E-TESS (Table 5), fairly recognizing all the six emotions, even though the emotions, "*disgust*" and "*fear*" are expressed differently in English and Bangla languages when expressed by their native speakers. This outcome is conclusive because, to reiterate, both BESS and EESS were developed by native Bangla speakers, and from the prior experiment, we deduced that non-native speakers might convey emotions as if they were expressing themselves in their native language. Hence, it is intuitive that this combination of datasets (BESS as training, EESS as testing) will result in a reasonable recognition rate. The result of this experiment also reinforces our claim that the non-native speakers tend to express their emotions in English, likewise their native tongue. These claims are further corroborated in the following sections.

### 5.3.1 Binary Classification of Emotions

This experiment includes substantiating the claims formulated for the emotions, "*disgust*" and "*fear*", which were found to have different expressions in Bangla and English languages when spoken by their native speakers. The third experiment is replicated in this instance, which involved training the SVM classifier by one speech set and separately testing it with the remaining two. However, instead of multi-class classification, binary classification for each of the six emotions is carried out. For example, considering the emotion, *happy*, any instance of any dataset, labelled as either *disgust* or *fear*, or any of the remaining emotions, can be equivalently labelled as, *not happy*. In this manner, for each dataset, there will be only two emotions, for example, *"happy"* – *"not happy"*, *"angry"* – *"not angry"*, *"neutral"* – *"not neutral"*, *"sad"* – *"not sad"*, *"disgust"* – *"not disgust"*, and *"fear"* – *"not fear"*.

Each of the training sets contains an equal number of labelled training samples for the binary emotions. For instance, the emotion, *happy*, has 50 audio samples and the emotion, *not happy*, also includes 50 audio samples (5 remaining emotions × 10 audio samples per emotion). This procedure is replicated for the rest of the emotions for each dataset. After this, the SVM model is trained on the binary labelled speech set and tested with a different binary labelled speech set for the same emotion. Testing was conducted with subsets of 30 audio samples from the testing dataset, and the overall recognition rate was recorded.

As delineated in Table 6 and Table 7, the results from this binary classification further substantiate and validate the claims purported in experiment3. The emotions, *disgust* and *fear* were completely un- recognized with BESS as the training and EESS as the testing datasets. Furthermore, with E-TESS as the training and either of the two remaining speech sets (BESS and EESS) as the testing datasets, the emotions *disgust* and *fear,* were left unrecognized as well. The confusion matrix for the "*disgust*" emotion (with E-TESS as the training and EESS as the testing datasets), as shown in Table 8, demonstrates that no true positives were recorded for this emotion, in which case the *precision* and the *recall* of the model becomes 0. Consequently, the *F1-score*, a weighted average of *precision* and *recall*, will be 0 as well. Therefore, it may be concluded that the emotions *disgust* and *fear* are expressed differently in Bangla and English languages. In contrast, these emotions were fairly recognized with BESS and EESS as the training and the testing datasets, respectively, which corroborates the claim that non-native speakers of a language tend to express their emotions, likewise their native tongue.

Table 6: Experiment 3.1 – Recognition rate of SVM in Binary Classification of the emotions – *happy*, *angry,* and *neutral*.

| Training Dataset | Testing Dataset | Binary Classification Emotion Performance (%) | | | | | | | | |
|---|---|---|---|---|---|---|---|---|---|---|
| | | happy | not happy | overall | angry | not angry | overall | neutral | not neutral | overall |
| Bangla (BESS) | English (EESS) | 53 | 93 | 88.33 | 77 | 94 | 90 | 86 | 97 | 95 |
| | English TESS (E-TESS) | 71 | 91 | 86.67 | 78 | 96 | 93.33 | 100 | 100 | 100 |
| English TESS (E-TESS) | English (EESS) | 52 | 89 | 81.67 | 74 | 92 | 88.33 | 60 | 92 | 86.67 |
| | Bangla (BESS) | 53 | 91 | 85 | 76 | 95 | 91.67 | 53 | 91 | 85 |

Table 7: Experiment 3.1 – Recognition rate of SVM in Binary Classification of the emotions – *sad*, *disgust,* and *fear*.

| Training Dataset | Testing Dataset | Binary Classification Emotion Performance (%) | | | | | | | | |
|---|---|---|---|---|---|---|---|---|---|---|
| | | sad | not sad | overall | disgust | not disgust | overall | fear | not fear | overall |
| Bangla (BESS) | English (EESS) | 86 | 96 | 93.33 | 57 | 94 | 90 | 77 | 97 | 95 |
| | English TESS (E-TESS) | 60 | 90 | 91.67 | **0** | 90 | 81.67 | **0** | 79 | 65 |
| English TESS (E-TESS) | English (EESS) | 17 | 79 | 66.67 | **0** | 89 | 80 | **0** | 92 | 85 |
| | Bangla (BESS) | 15 | 83 | 70 | **0** | 86 | 75 | **0** | 88 | 78.33 |

Table 8: Confusion Matrix for Disgust Emotion (Training Dataset: English TESS, Testing Dataset: English).

| | | Predicted Label | |
|---|---|---|---|
| | | Disgust | Not Disgust |
| True Label | Disgust | 0 | 4 |
| | Not Disgust | 2 | 24 |

Across all the experiments, the emotional states: *happy, angry, neutral*, and *sad* were identified regardless of language and native/non-native speakers. However, the emotions: *disgust* and *fear* brought about some discrepancies. Considering the two hypotheses formed in section 3.1, from these suites of experiments, it can be inferred that for Bangla and English languages, the emotional state *happy* is perhaps language and speaker (native/non-native) independent while the emotional states, *disgust* and *fear* are perhaps language and speaker (native/non-native) dependent. Additionally, non-native speakers are found to convey emotions analogous to their expressions in their native language.

## 6 Conclusion

In this study, we developed 3 datasets namely, Bangla Emotional Speech Set (BESS), English Emotional Speech Set (EESS), and English TESS (E-TESS) to evaluate the language independence of Speech Emotion Recognition (SER), featuring 6 emotions, such as - *happy*, *angry*, *neutral*, *sad*, *disgust*, and *fear* [1], [2], [15], [20], [23], [31], in English and Bangla languages through language-independent vocal feature selection. The datasets BESS and EESS were developed by 11 native Bangla speakers and E-TESS was developed as a subset of the Toronto Emotional Speech Set (TESS) while preserving the speech intelligibility functions across the datasets.

We coordinated 3 experiments, where we deployed SVM for classifying emotions, as it is one of the most widely used classifiers for SER systems. Although the performance of the model varied across the 3 experiments, the reliability of recognizing different emotions was convincing. The results from the first experiment revealed that native speakers tend to express their emotions better in their native language than non-native speakers of that language. From the second and the third experiments, it was observed that non-native speakers of a language have a keen proclivity of expressing their emotions, likewise their native language. Nonetheless, these experiments further point out that there may be certain differences in languages that govern the expression of different emotions.

This claim was eventually solidified by the end of the third experiment when the emotional states, *disgust* and *fear* were revealed to be language and speaker dependent. The factors behind this contrast may be credited to cultural, background, environmental or communal differences. However, this study also demonstrated that the emotional states such as *happy*, *angry, neutral*, and *sad* were moderately recognized irrespective of language and native/non-native speakers. Therefore, we can deduce that there may be language-specific differences for certain, if not all emotions.

The findings direct us towards the conclusion that SER in Bangla and English is mostly language independent. However, some disparity exists in the emotional states owing to cultural, environmental, or communal factors. Hence, we hope that the community will find the results of this study applicable towards achieving comprehensive control over SER regardless of language. In the future, we intend to extend our analysis of language independence of SER by including more languages, speakers, and language independent features.

## Acknowledgments

The authors express their heartfelt gratitude to the participants for their valuable time and effort for making this study possible.

## Declaration Of Interests

The authors do not declare any potential conflict of interest that may alter the outcomes of this study in any manner and approve this version of the manuscript for publication.